\documentclass[twoside,leqno,twocolumn]{article}
\usepackage{ltexpprt}
\usepackage{color}
\usepackage{amsmath}
\usepackage{amssymb}
\usepackage{hyperref}
\usepackage{sidecap}
\usepackage{graphicx}
\begin{document}

\title{\Large Knowledge Transfer with Medical Language Embeddings\thanks{Supported by the Memorial Hospital and the Sloan Kettering Institute (MSKCC; to G.R.). Additional support for S.L.H. was provided by the Tri-Institutional Training Program in Computational Biology and Medicine.}}

\author{Stephanie L. Hyland\thanks{Tri-Institutional Training Program in Computational Biology and Medicine, Weill Cornell Medical College}\hspace{0.15cm}\textsuperscript{\ddag}
\and
Theofanis Karaletsos\footnote{Computational Biology Program, Memorial Sloan Kettering Cancer Center (MSKCC)}
\and
Gunnar R\"{a}tsch\footnotemark[\value{footnote}]}
\date{}

\maketitle


\begin{abstract} \small\baselineskip=9pt
Identifying relationships between concepts is a key aspect of scientific
knowledge synthesis. Finding these links often requires a researcher to
laboriously search through scientific papers and databases, as the size of
these resources grows ever larger. In this paper we describe how distributional
semantics can be used to unify structured knowledge graphs with unstructured
text to predict new relationships between medical concepts, using a
probabilistic generative model. Our approach is also designed to ameliorate 
data sparsity and scarcity issues in the medical domain, which make language 
modelling more challenging. Specifically, we integrate the medical relational database (SemMedDB) with text from electronic health records (EHRs) to perform 
knowledge graph completion. We further demonstrate the ability of our model to predict relationships between tokens not appearing in the relational database.
\end{abstract}

\section{Introduction}
The accelerating pace of scientific progress presents both challenge and opportunity to researchers and healthcare providers. Reading and comprehending the ever-growing body of literature is a difficult but necessary part of knowledge discovery and synthesis. This is particularly important for biomedical research, where therapeutic breakthroughs may rely on insights derived from disparate subfields. Curating literature at such breadth and scale is infeasible for individuals, necessitating the development of domain-specific computational approaches.

We present here a method using \emph{language embeddings}. Such an embedding is a representation of the tokens of a language (such as words, or objects in a controlled vocabulary) as elements of a vector space. Semantic similarity is then captured by vector similarity, typically through Euclidean or cosine distance. The dimensionality of the space is typically much less than the size of the vocabulary, so this procedure allows tokens to be represented more compactly while also capturing semantics. Such representations can be used as features in downstream language-processing tasks. In our case, we aim to exploit the embedding \textit{itself} to discover new relationships between tokens. This is possible because our embedding procedure defines a probability distribution over token-relationship-token triples, allowing for questions such as `is \texttt{abdominal pain} more likely to be \texttt{associated with} \texttt{acute appendicitis} or \texttt{pulmonary tuberculosis}?', or `how is \texttt{radium} related to \texttt{carcinoma}?'\footnote{These are real examples from \texttt{SemMedDB}.}

The tokens of interest are chiefly Concept Unique Identifiers (CUIs) from the Unified Medical Langauage System (UMLS) \cite{bodenreider2004unified}. These represent discrete \emph{medical concepts}, which may require several words to describe, for example: \texttt{C0023473: chronic myelogenous leukemia}. We consider it more meaningful and interesting to consider relationships between CUIs rather than words themselves, when possible. We exploit the exisence of \texttt{SemMedDB} \cite{Kilicoglu01122012}, a database of semantic predications in the form of subject-relationship-object triples, where the subjects and objects are such CUIs. These were derived from PubMed abstracts using the tool SemRep \cite{Rindflesch2003462}. We combine this structured data with unstructured text consisting of clinical notes written by physicians at Memorial Sloan Kettering Cancer Center (MSKCC).

\section{Related Work}
Neural language models \cite{Bengio:2003:NPL:944919.944966} are an approach to learning embeddings which use a word's representation to \emph{predict} its surrounding context. This relies on the fact that words with similar meanings have similar contexts (the distributional hypothesis of language \cite{sahlgren2008distributional}), which forces their representations to be similar. Intriguingly, it was observed \cite{mikolov2013distributed} \cite{bordes2013translating} that the geometry of the resulting space preserved \emph{functional relationships} between terms. An example is a consistent offset vector existing between `Berlin' and `Germany', and `Dublin' and `Ireland', seemingly representing the relationship \texttt{capital city of country}. This property has been exploited to perform knowledge-base completion, for example \cite{bordes2013translating} \cite{socher2013reasoning} \cite{weston2013connecting}, however these approaches have restricted their attention to edge-discovery \emph{within} a knowledge graph. To \emph{extend} such a graph we therefore developed a model \cite{hyland2016} which can combine structured and unstructured data sources while explicitly modelling the types of relationships present in the structured data.

Despite the popularity of language embeddings in the broader natural language processing (NLP) community, the biomedical domain has yet to fully exploit them. Pedersen et \emph{al.} \cite{Pedersen2007288} highlight the need to perform \emph{domain-specific} NLP and discuss measures of semantic relatedness. Other recent applications include using representations of nominal elements of the EHR to predict hospital readmission \cite{Krompaß2014}, identifying adverse drug reactions \cite{henriksson_representing_2015}, and clinical concept extraction \cite{jonnalagadda2012enhancing}.

\section{Approach}
\subsection{Model}
We briefly describe the \texttt{bf} model; see our earlier paper \cite{hyland2016} for more details. This is a probabilistic generative model over directed subject-relationship-object triples ($S$, $R$, $O$). Subject and object are both tokens from the vocabulary (e.g., UMLS CUIs), although following \cite{mikolov2013efficient} and \cite{goldberg2014word2vec} we give them independent representations. This is formulated mathematically through an energy function,
\begin{equation}
\mathcal{E}(S,R,O|\Theta) = -\frac{\mathbf{v}_O\cdot G_R \mathbf{c}_S}{\|\mathbf{v}_O\|\| G_R \mathbf{c}_S\|}
\label{eq:energyfunction}
\end{equation}
Entities $S$ and $O$ are represented as vectors, while each representation $R$ corresponds to an \emph{affine transformation} on the vector space. Intuitively, our energy function is the cosine distance between (the representation of) $O$ and $S$ \emph{under the context of $R$}, where this context-specific similarity is achieved by first transforming the representation of $S$ by the affine transformation associated to $R$.

This energy function defines a Boltzmann probability distribution over $(S, R, O)$ triples,
\begin{equation}
P(S,R,O|\Theta) = \frac{1}{Z(\Theta)}e^{-\mathcal{E}(S,R,O|\Theta)}
\label{eq:pdf}
\end{equation}
where the denominator is the partition function, $Z(\Theta) = \sum_{s,r,o}e^{-\mathcal{E}(s,r,o|\Theta)}$. Equation~\ref{eq:pdf} defines the probability of observing a triple $(S, R, O)$, given the embedding $\Theta$, which is the set of all vectors $\{\mathbf{c}_s$, $\mathbf{v}_o\}_{s, o \in \textrm{tokens}}$ and matrices $\{G_r\}_{r \in \textrm{relationships}}$.

\subsection{Training}
To learn the embedding (the parameters $\Theta$ consisting of all word vectors $\mathbf{c}_s$, $\mathbf{v}_o$, and the relationship matrices $G_r$), we maximise the joint probability of a set of \emph{true} triples $(S, R, T)$ under this model. Likely pairs have a high cosine similarity (low energy) in the context of their shared relationship, requiring similar vector representations. We employ stochastic maximum likelihood for learning, approximating gradients of the partition function using persistent contrastive divergence \cite{tieleman2008training}.

In all cases, we perform early stopping using a held-out validation set. The hyperparameters of the model are as follows: vector dimension is 100, batch size is 100, we use 3 rounds of Gibbs sampling to get model samples, of which we maintain one persistent Markov chain. The learning rate is $0.001$ and we use a $l_2$ regulariser with strength 0.01 on $G_r$ parameters. To make learning more stable, we use Adam \cite{kingma2014adam} with hyperparameters as suggested in the original paper.

\subsection{Prediction}
Equation~\ref{eq:pdf} defines a joint distribution over triples. However, we are often interested in \emph{conditional} probabilities: given a pair of entities $S$ and $O$, which $R$ most likely exists between them (if any)? Such a distribution over $R$ (or equivalently $S$, $O$) can easily be derived from the joint distribution, for example:
\begin{equation} 
P(R | S, O; \Theta)  =\frac{e^{-\mathcal{E}(S, R, O|\Theta)}}{\sum_{r}e^{-\mathcal{E}(S, r, O|\Theta)}}
\label{eq:cond}
\end{equation}
The cost of calculating the conditional probability is at worst linear in the size of the vocabulary, as the (generally intractable) partition function is not required.

\section{Experiments}
\subsection{Data preparation}
We train the model on two types of data: \emph{unstructured} (\texttt{EHR}) and \emph{structured} (\texttt{SemMedDB}). The unstructured data is a corpus of de-identified clinical notes written by physicians at MSKCC. We process raw text by replacing numbers with generic tokens such as \texttt{HEIGHT} or \texttt{YEAR}, and removing most punctuation. In total, the corpus contains 99,334,543 sentences, of which 46,242,167 are unique. This demonstrates the prevalence of terse language and sentence fragments in clinical text; for example the fragment \texttt{no known drug allergies} appears 192,334 times as a sentence. We identify CUIs in this text by greedily matching against strings associated with CUIs (each CUI can have multiple such strings). This results in 45,402 unique CUIs, leaving 270,100 non-CUI word tokens. We note that the MetaMap \cite{aronson2001effective} tool is a more sophisticated approach for this task, but found it too inefficient to use on a dataset of our size. To generate $(S, R, O)$ triples, we consider two words in a \texttt{appears in a sentence with} relationship if they are within a five-word window of each other.

The structured data (\texttt{SemMedDB}) consists of \texttt{CUI}-\texttt{relationship}-\texttt{CUI} statements, for example \texttt{C0027530(Neck)} is \texttt{LOCATION OF C0039979(Thoracic Duct)} or \texttt{C0013798(Electrocardiogram) DIAGNOSES C0026269(Mitral Valve Stenosis)}. These were derived from PubMed abstracts using SemRep \cite{Rindflesch2003462}. \texttt{SemMedDB} contains 82,239,653 such statements, of which 16,305,000 are unique. This covers 237,269 unique CUIs. 

Since the distribution of CUI/token frequencies has a long tail in both data sources, we threshold tokens by their frequency. Firstly, tokens (words of CUIs) must appear at least 100 times in either dataset, and then at least 50 times in the pruned datasets.
That is, in the first round we remove sentences (in \texttt{EHR}) or statements (in \texttt{SemMedDB}) containing `rare' tokens. In addition, the 58 relationships in \texttt{SemMedDB} also exhibit a long-tailed frequency distribution, so we retain only the top twenty.

From this pool of $(S, R, O)$ triples (from \texttt{EHR} and \texttt{SemMedDB}) we create fixed test sets (see next subsection) and smaller datasets with varying relative abundances of each data type, using 0, 10, 50, 100, 500, and 1000 thousand training examples. The final list of tokens has size $W=45,586$, with 21 relationships: twenty from \texttt{SemMedDB} and an additional \texttt{appears in sentence with} from \texttt{EHR}. Of the $W$ tokens, 7,510 appear in both data sources. These overlapping tokens are critical to ensure embeddings derived from the knowledge graph are consistent with those derived from the free text, allowing information transfer.

\subsection{Knowledge-base completion}
\paragraph{Experimental design}
As the model defines conditional distributions for each element of a triple given the remaining two (Equation~\ref{eq:cond}), we can test the ability to predict new components of a knowledge graph. For example, by selecting the best $R$ given $S$ and $O$, we predict the relationship (the type of edge) between tokens $S$ and $O$. 

Without loss of generality, we describe the procedure for generating the test set for the $R$ task. We select a random set of $S, O$ pairs appearing in the data. For each pair, we record all entities $r$ which appear in a triple with them, removing these triples from the training set. The $S, O \rightarrow \{r_i\}_i$ task is then recorded in the test set. Evidently, there may be many correct completions of a triple; in this case we expect the model to distribute probability mass across all answers. How best to evaluate this is task-dependent; we consider both the \emph{rank} and the \emph{combined probability mass} in these experiments.

\begin{figure}
\includegraphics[scale=0.7]{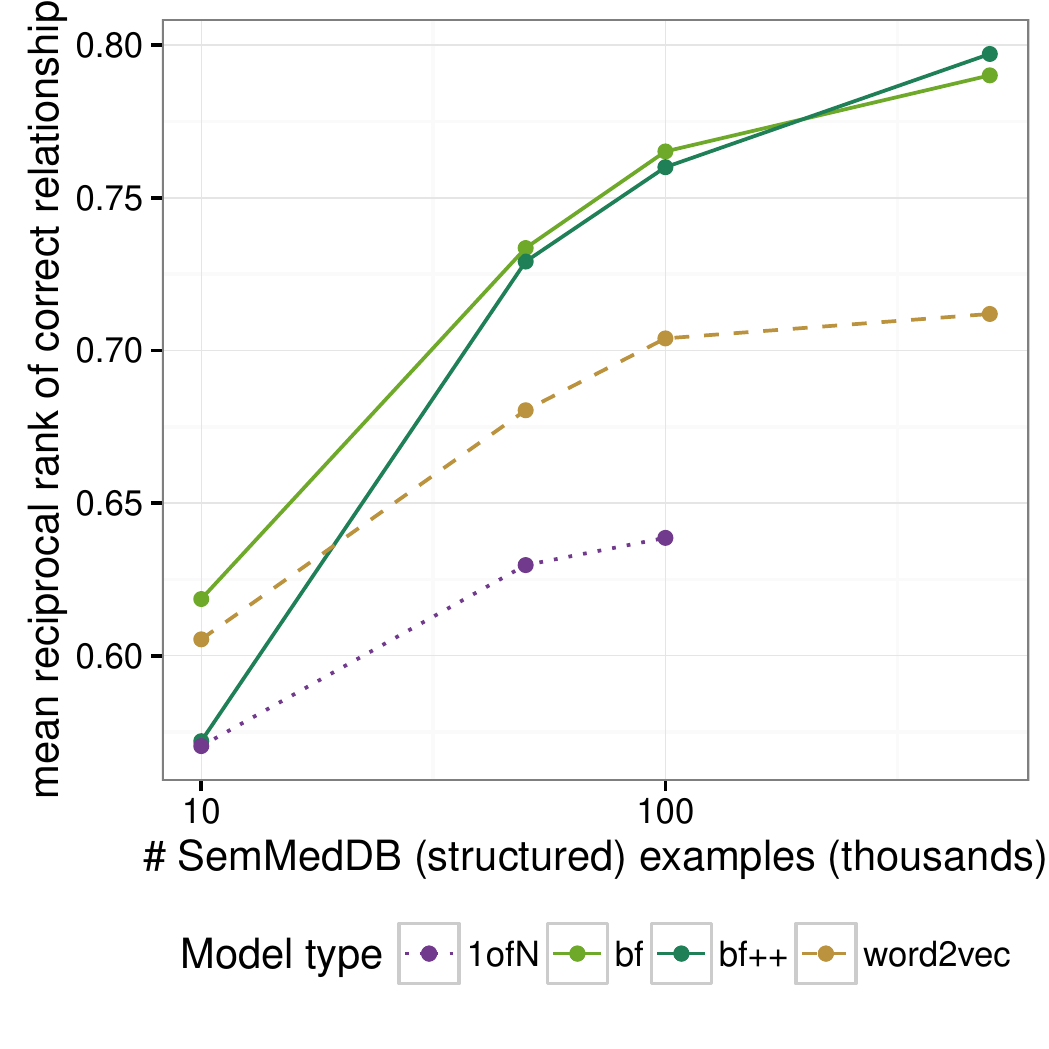}
\vspace{-0.5cm}
\caption{With more evidence from the knowledge graph, the model better predicts the correct relationship for a given ($S$, $O$) pair. \texttt{bf++} has an additional 100,000 triples from \texttt{EHR}: with little structured data, so much off-task information is harmful, but provides some benefit when there is enough signal from the knowledge graph. Baselines are a random forest taking $[f(S):f(O)]$ as an input to predict the label $R$, where the feature representation $f$ is either a 1-hot encoding (\texttt{1ofN}) or 200-dimensional \texttt{word2vec} vectors trained on PubMed. \texttt{1ofN} proved too computationally expensive for large data.}
\label{fig:nus}
\vspace{-0.5cm}
\end{figure}

\paragraph{Results}
Figure~\ref{fig:nus} shows results for the task of predicting $R$ given $S$ and $O$. The model produces a ranking of all possible $R$s (high probability $\rightarrow$ low rank) and we report the mean reciprocral rank of the \emph{lowest-ranked} correct answer over the test set. We use this metric to evaluate the utility of these predictions in \emph{prioritising} hypotheses to test: we would like \emph{any} correct answer to be ranked highly, and don't apply a penalty for a failure to capture alternative answers. Results for our model are marked by \texttt{bf}\footnote{\texttt{bf} stands for `br\'{i}-focal', which means \emph{word meaning} in Irish.} and \texttt{bf++}. The latter model uses an additional 100,000 training examples from the \texttt{EHR}: these are `off-task' information. As a baseline we consider a random forest trained to predict $R$ given the concatenation $[f(S):f(O)]$, where the representation $f$ is either: a) \texttt{1ofN}: each token has a binary vector of length $W$ ($W$ = 45,586), b) \texttt{word2vec}: each token has a 200-dimensional vector obtained by running \texttt{word2vec} \cite{mikolov2013efficient} trained on PubMed \cite{moendistributional}. We note that the PubMed corpus contains over 2 billions tokens, far more data than was available to \texttt{bf}. We additionally trained \texttt{TransE} \cite{bordes2013translating} on this data, but it proved unsuited to the task (data not shown).

As we can see, adding examples from \texttt{SemMedDB} improves performance for all model types, but \texttt{bf} seems to make better use of the additional data. In spite of its very large input vector size ($2W = 91172$), \texttt{1ofN} struggles, likely as it treats all tokens as independent entities. We note that for \texttt{bf++}, performance is \emph{degraded} when the amount of structured data is low. This is consistent with earlier observations on non-medical data \cite{hyland2016}, as the quantity of `off-task' information added is in this case comparable to that of `on-task'. Interestingly however, the model appears slightly \emph{better able} to exploit more structured data when some `semantic background' is provided by \texttt{EHR}.

\subsection{Information transfer}
\paragraph{Experimental design}
As mentioned, the model is capable of combining structured and unstructured data. In \cite{hyland2016} we observed that classification performance on a knowledge base could be improved by addition of unstructured data. However, the task in that case was quite `easy'; the model simply needed to differentiate between true and false triples. Here we consider the harder problem of correctly selecting \emph{which} entity would complete the triple. 

In addition to possibly improving performance, access to unstructured data provides the opportunity to \emph{augment} the knowlede base. That is, we can predict relationships for tokens \emph{not appearing} in \texttt{SemMedDB}. This uses the joint embedding of all tokens into one vector space, regardless of their data source. The geometric action of the relationships learned from \texttt{SemMedDB} can then be applied to the representation of any token, such as those uniquely found in \texttt{EHR}. We note that this procedure amounts to \emph{label transfer} from structured to unstructured examples, which can be understood as a form of semi-supervised learning.

To generate ground truth for this task, we select some tokens $\{T_i\}$ (these could appear as $S$ or $O$ entities) found in both \texttt{SemMedDB} and \texttt{EHR} and remove them from \texttt{SemMedDB}, recording them to use in the test set. Put another way, as in the previous setting, during the `random' selection of $S, O$ (still wlog) pairs, we make sure all of these recording them to use in the test set. Put another way, as in the previous setting, during the `random' selection of $S, O$ (still wlog) pairs, we make sure all $T_i$ in the deletion list are included, alongside any other tokens which appear in a \texttt{SemMedDB}-derived relationship with them. The task is then to use purely \emph{semantic} similarity gleaned from \texttt{EHR} to place these tokens in the embedding space such that the action of relationship operators is still meaningful. 

\begin{figure}
\includegraphics[scale=0.6]{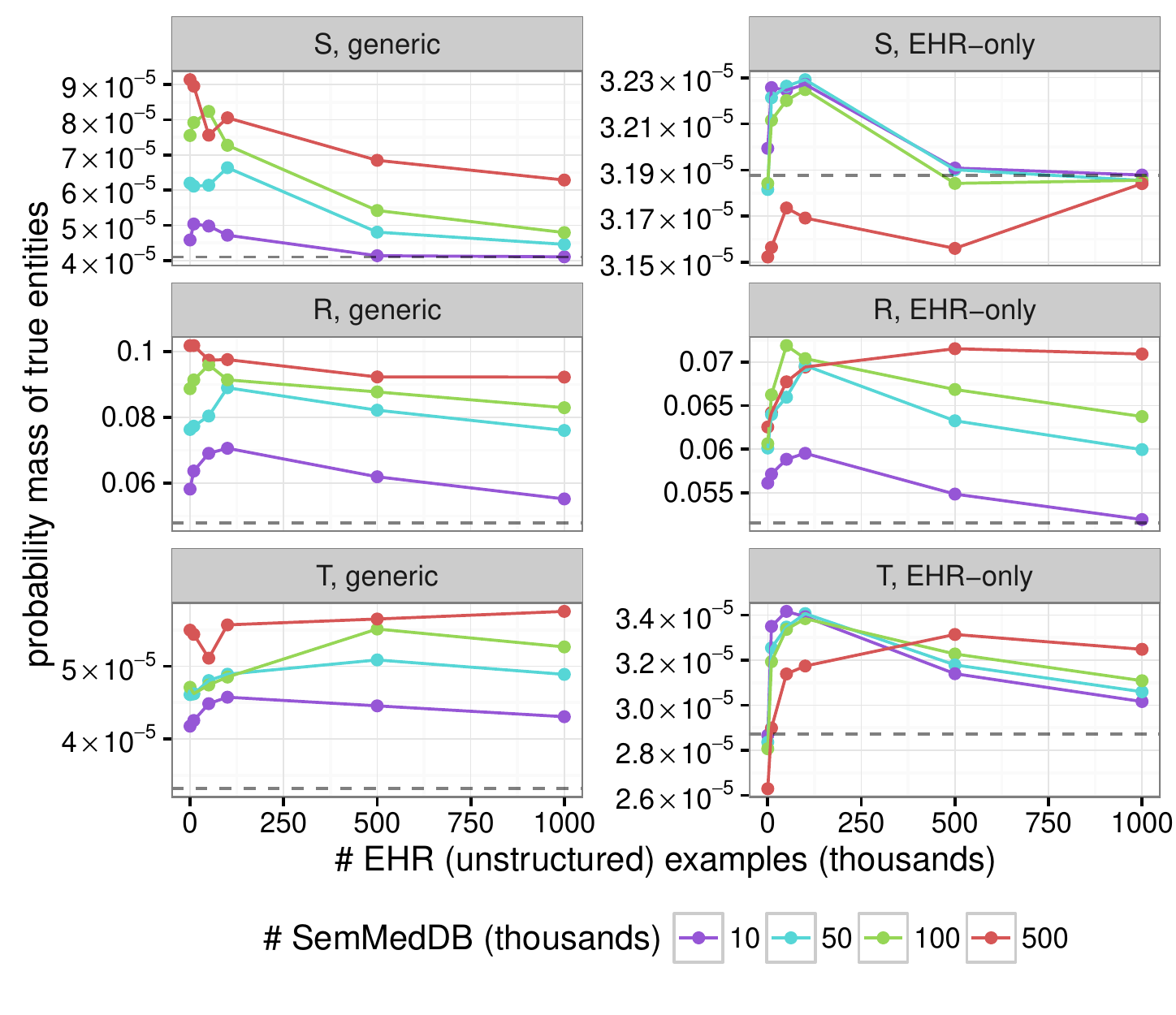}
\vspace{-0.9cm}
\caption{Total probability mass assigned to correct answers for all tasks. The right column shows results for test triples where at least one of $S$ and $O$ is found \emph{only} in \texttt{EHR}, and therefore represents the \emph{knowledge transfer} setting. Information about relationships found in \texttt{SemMedDB} must be transferred through the joint embedding to enable these predictions. Grey dotted lines represent a random-guessing baseline.}
\vspace{-0.4cm}
\label{fig:sumprob}
\end{figure}

\paragraph{Results}
Figure~\ref{fig:sumprob} shows results on all three tasks (predicting $S$, $R$, $O$ given the remaining two), as a function of the \emph{type of test example}. The right column of results is for test entities involving at least one element \emph{not appearing} in \texttt{SemMedDB}. As we are now interested in the \emph{embeddings themselves} we report the probability mass of true entities, feeling this better captures the information contained in the embeddings. That is, it is no longer sufficient for the model to correctly predict \emph{a single} answer, we want it to assign appropriate probability mass to \emph{all} correct answers. The dotted grey lines demonstrate the random baseline, where all tokens are equally likely. The probability mass assigned by the baseline is therefore equal to $k/W$ (or $k/R$) where $k$ is the average number of correct options in that task type.

There are several observations to be made here:
\begin{itemize}
\item Most of the time, performance is best with a non-zero, but \emph{relatively small} amount of \texttt{EHR} data ($x$-axis). This supports our observations that off-task information improves embeddings, but can `drown out' signal if it dominates relative to the on-task examples. This can be improved by including a pre-factor on gradient contributions from the off-task data to adjust their contribution relative to the structured examples, as demonstrated in our previous work \cite{hyland2016}.
\item The \texttt{EHR-only} setting is much harder, as anticipated. In the case of $S$ and $O$ it is comparable to the random baseline. For $R$ however, the model successfully assigns probability mass when there is enough \texttt{SemMedDB} data available.
\item The $S$ and $O$ tasks are not symmetric. The $S$ task features slightly more correct options on average than $O$ (1.87 and 1.5 respectively, for the \texttt{generic} task), but this does not account for the difference in proportional performance relative baseline, especially at low \texttt{EHR} abundance. A possible explanation is the energy function (Equation~\ref{eq:energyfunction}): it does not treat $S$-type and $O$-type variables identically. However, experiments using the Frobenius norm of $G_R$ in the denominator of $\mathcal{E}$ did not remove asymmetry, so it is likely that the tasks are simply not equivalent. This could arise due to bias in the directionality of edges in the knowledge graph.
\end{itemize}

We conclude that it is possible to use the joint embedding procedure to predict $R$ for pairs of $S$, $O$ entities even if they do not appear in \texttt{SemMedDB}. For the harder $S$ and $O$ tasks, the model generally succeeds in improving visibly over baseline, but its assignments are still quite `soft'. This may reflect premature stopping during training (most results reported were before 50 epochs had elapsed), an insufficiently powerful model formulation, or an excess of noise in the training data. Many predicates in \texttt{SemMedDB} are vague, and some relationships lend themselves to a one-to-many situation, for example \texttt{part of}, or \texttt{location of}. A core assumption in our model is that a token with fixed vector representation can be transformed by a single affine transformation to be similar to its partner in a relationship. Many-to-one (or vice-versa) type relationships requires that multiple unique locations must be mapped to the same point, which necessitates a rank-deficient linear operator or a more complex transformation function (one which is locally-sensitive, for example). Future work in relational modelling must carefully address the issue of many-to-many and hierarchical relationships.

\section{Discussion}
Distributed language representations have seen limited application in healthcare to date, but present a potentially very powerful tool for analysis and discovery. We have demonstrated their use in knowledge synthesis and text mining using a probabilistic generative model which combines structured and unstructured data. These embeddings can further be used in downstream tasks, for example to reduce variation in language use between doctors (by identifying and collapsing similar terms), for `fuzzy' term-matching, or as inputs to \emph{compositional} approaches to represent larger structures such as sentences, documents, or even patients. Expressive knowledge representations such as these will be facilitate richer clinical data analysis in the future.

\newpage
\bibliography{reffos}

\end{document}